\documentclass[runningheads]{llncs}
\usepackage[T1]{fontenc}
\usepackage{graphicx}
\usepackage{amsmath}
\usepackage{color}
\usepackage{multirow}
\usepackage{arydshln}
\usepackage{float}
\usepackage{amssymb}
\usepackage{cleveref}
\usepackage{cite}
\begin{document}
\title{Enhancing Frequency for Single Image Super-Resolution with Learnable Separable Kernels}
%
%
\author{Heng Tian\inst{1, 2}\orcidID{0000-0003-2505-6871}}
\authorrunning{F. Author et al.}
%
\institute{Key Laboratory of Smart Manufacturing in Energy Chemical Process, Ministry of Education, East China
University of Science and Technology, Shanghai, 200237, China.  \and
Department of Computer Science and Engineering, East China
 University of Science and Technology, Shanghai, 200237, China.\\
\email{tianheng@mail.ecust.edu.cn}}
\maketitle              
\begin{abstract}
Existing approaches often enhance the performance of single-image super-resolution (SISR) methods by incorporating auxiliary structures, such as specialized loss functions, to indirectly boost the quality of low-resolution images. In this paper, we propose a plug-and-play module called Learnable Separable Kernels (LSKs), which are formally rank-one matrices designed to directly enhance image frequency components. We begin by explaining why LSKs are particularly suitable for SISR tasks from a frequency perspective. Baseline methods incorporating LSKs demonstrate a significant reduction of over 60\% in both the number of parameters and computational requirements. This reduction is achieved through the decomposition of LSKs into orthogonal and mergeable one-dimensional kernels. Additionally, we perform an interpretable analysis of the feature maps generated by LSKs. Visualization results reveal the capability of LSKs to enhance image frequency components effectively. Extensive experiments show that incorporating LSKs not only reduces the number of parameters and computational load but also improves overall model performance. Moreover, these experiments demonstrate that models utilizing LSKs exhibit superior performance, particularly as the upscaling factor increases.

\end{abstract}
\section{Introduction}\label{sec:introduction}
    In real-world scenarios, high-quality images with clear and accurate object representations are essential for both human interpretation and machine processing to make informed decisions. However, low-quality images are frequently encountered due to poor lighting conditions or limitations in imaging equipment. To address this issue, Single Image Super-Resolution (SISR)~\cite{medical_sr, img_restore} techniques are employed to reconstruct high-resolution (HR) images from low-resolution (LR) counterparts that lack detail and appear blurry.

    Since the introduction of the first CNN-based method, SRCNN~\cite{SRCNN}, numerous CNN-based approaches~\cite{ESPCN, VDSR, EFDN, ECB} have been proposed to tackle SISR tasks. These approaches, which often involve deeper and more complex networks, have demonstrated superior performance compared to classical methods~\cite{cm1, cm2}. However, deploying these large SISR models on resource-constrained platforms, such as mobile devices or FPGAs, while maintaining real-time performance and low parameter counts, presents a significant challenge. Generally, deeper neural networks achieve better performance but require more memory resources~\cite{google-v3}.

    The primary difference between LR images and their HR counterparts is the absence of high-frequency information, which is crucial for capturing fine details. Since edge information is a key frequency feature in image processing, operators like Laplace or Sobel are commonly used for edge feature extraction or enhancement. Therefore, enhancing frequency components is a critical research direction~\cite{EFDN, FAD} for SISR tasks, and various edge-related operators have been utilized in previous work~\cite{ECB}. However, these specialized and non-learnable operators are often employed as auxiliary tools within the primary framework.

    The kernel mostly used in CNNs is a square kernel with the same size in length and width, such as $3\times3$ or $5\times 5$. However, these square kernels find it difficult to effectively enhance frequencies when using generic loss functions like L1 or L2 loss. To address this, Gradient Variance (GV) loss~\cite{gvloss} has been proposed to recover sharp edges. Nonetheless, this approach is indirect and balancing it with other loss functions can be challenging. 
    
\begin{figure}
        \begin{center}
            \includegraphics[width=0.4\linewidth]{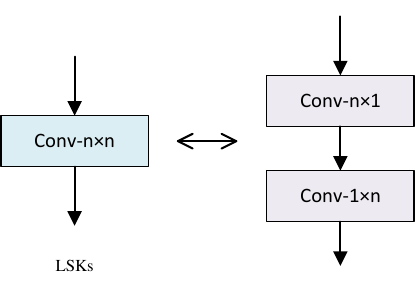}
        \end{center}
        \caption{ Lsks can be decomposed into two 1D kernels, which improves the efficiency and reduces the number of parameters. It is worth noting that LSKs are different from Inception-v3~\cite{google-v3}, which replaces any $n \times n $ convolutions by a $1 \times n $ convolutions followed by a $n \times 1 $ convolutions.}
        \label{fig:lsk}
    \end{figure}

    In this paper, we propose Learnable Separable Kernels (LSKs) as a substitute for traditional square kernels, enabling the model to actively learn and enhance frequency representations. A matrix with rank $N$ can be decomposed into the sum of $N$ matrices with rank $1$. We refer to these matrices with rank $1$ separable matrices (kernels), which can be decomposed into two 1D multiplied kernels (Fig.~\ref{fig:lsk}). From a mathematical perspective, special edge-related rank-1 kernels such as Sobel have the ability to increase frequency. Therefore, we explicitly introduce the LSKs rather than fixed kernels into the SISR to enhance the frequency. To reduce the number of parameters and computation, we decompose LSKs into 1D kernels during training and inference.
    
    %
    Mathematically, the entire process that replacing square kernels with 1D kernels is equivalent to decomposing the square kernels into the sum of a series of LSKs. Since the absence of nonlinear transformations (activation), we can merge the two-layer 1D kernels into the separable kernels. Therefore, this replacement does not significantly reduce the representation ability of the model, instead, the model gains the ability to enhance the frequency and improve the performance on the SISR task.
    

    The LR images are typically obtained by downsampling the original images at different scales. As the scale increases, the more high-frequency information LR loses. In such cases, our experiments show that LSKs perform better than normal kernels in large-scale down-sampling, which is consistent with the analysis that LSKs can increase frequency. In summary, our main contributions can be summarized as follows:
    \begin{itemize}
        \item
        Based on the specificity of the SISR task, we introduce the LSKs into the SISR task to enhance frequency combined with the assistance of neural network learning.
        \item
        We demonstrate the ability of LSKs from the perspective of interpretability and find that LSKs are well-suited for SISR tasks.
        \item
        We apply the LSKs to three classical baseline methods and conduct comprehensive experiments on five benchmark datasets. The results demonstrate that the separable-kernel-version baseline methods are superior both in parameters and efficiency. Moreover, with the increase of scale factor in the SISR task, these separable version methods perform better than normal.
    \end{itemize}

\section{Related Work}\label{sec:related-work}

    \subsection{Single-image super-resolution}

    Unlike classification problems, Single Image Super-Resolution (SISR) methods generate high-resolution (HR) images from low-resolution (LR) inputs. Thus, effectively upsampling LR images to match the HR size is a key challenge. Typically, there are two frameworks for upsampling in SISR models: one involves upsampling the LR images to coarse HR images before applying SISR methods, and the other involves upsampling the images to the desired HR size after applying SISR methods~\cite{sr_survey}. Consequently, we can broadly categorize SISR methods into two types: pre-upsampling~\cite{SRCNN, VDSR} and post-upsampling~\cite{ESPCN, EFDN}.

    The first CNN-based SISR method, SRCNN~\cite{SRCNN}, adopts the pre-upsampling scheme. This approach has since become the most popular framework. In SISR, the main computational load comes from CNN operations, which are directly related to the image size. To improve computational efficiency, ESPCN~\cite{ESPCN} was proposed to reduce computational complexity by performing upsampling in the final step, allowing convolution operations to act on the smaller LR images rather than the larger HR images. This post-upsampling approach has also gained popularity. Since the introduction of ResNet~\cite{ResNet}, which facilitates deeper models by learning residuals, ResNet structures have become an essential component in SISR~\cite{VDSR, EFDN}.

    Most current SISR methods are based on these three primary structures. In this paper, we employ Learnable Separable Kernels (LSKs) in three baseline methods to evaluate performance: SRCNN (pre-upsampling), ESPCN (post-upsampling), and VDSR (ResNet-based). This allows us to comprehensively assess the effectiveness of LSKs across different SISR frameworks.


    \subsection{Frequency-enhanced methods}
    LR images exhibit a noticeable difference compared to their HR counterparts, primarily due to a lack of detail or high-frequency information. Consequently, improving or enhancing frequency components is a prominent research direction in SISR~\cite{EFDN, FAD}. The most relevant features to frequency information are edges, and many edge-related operators have been utilized in previous work~\cite{EFDN, ECB}. Notably, many edge-related operators, such as Sobel and Prewitt, are separable kernels. However, most previous models use non-learnable, specific operators, which limit the range of frequency features that can be captured. In this paper, we propose directly training models with LSKs, enabling the network to learn a broader range of frequency-enhanced kernels. This approach allows for the dynamic adaptation and optimization of frequency features, potentially leading to improved performance in SISR tasks.


    \subsection{Lightweight and Efficient methods}\label{subsec:kernel-optimization-based-methods}
    Reducing parameters~\cite{EFDN, ACB} or reducing the Floating point operations (Flops)~\cite{FAD} of neural networks are direct ways to obtain lightweight and efficient models. Considering the characteristic that the whole neural network is mainly composed of convolutional layers in CNN-based SISR models, the optimization direction mainly focuses on the convolution kernel. There are two natural strategies to compress the neural network. One is from the perspective of pruning: 1) pruning the convolution kernel or layer~\cite{kernelPruning, pruning_layer} directly, and the other is from the optimization: 2) reducing the number of parameters of the convolution kernel~\cite{low_rank1, low_rank2}.

    Due to the particularity of the CNN-based SISR methods, almost all calculations are convolution operations except activation. So, kernel-based optimization is a natural direction. To improve the accuracy of the model without increasing the inference time, Ding \textit{et al.}~\cite{ACB} train paralleled 1D asymmetric kernels to strengthen the square convolution kernels. Wang~\cite{EFDN} also parallels the RepVGG~\cite{RepVGG}, DBB~\cite{DBB}, and ECB~\cite{ECB} to improve performance in the stage of training. Different from the parallel scheme, we learn the series 1D asymmetric kernels in the stage of training.

    Optimization based on matrix decomposition~\cite{low_rank1, low_rank2} is another straightforward method. However, these works focus solely on reducing parameters and cannot improve performance. And they ignore the most important part: the separable kernels can enhance the frequency. Moreover, the mergeability of 1D kernels was not taken into consideration.

    \section{Method}\label{sec:methodology}

\begin{figure}
        \begin{center}
            \includegraphics[width=1\linewidth]{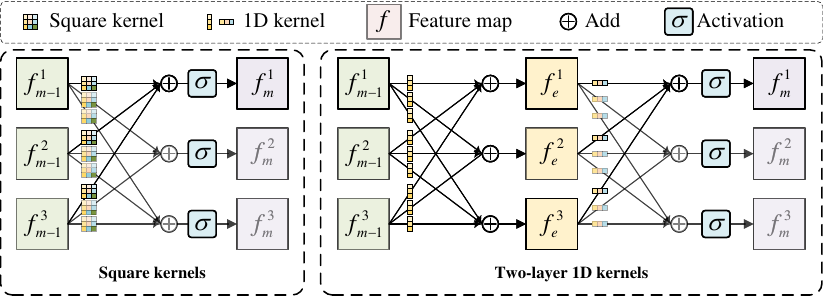}
        \end{center}
        \caption{The process of replacing the square kernels with LSKs and decomposing LSKs into 1D kernels. There will be an extra layer of feature maps, since the square kernels become two-layer 1D kernels. }
        \label{fig:cnn-p}
    \end{figure}

    Considering the specificity that images lose high-frequency information in the SISR task, we propose the LSKs instead of square kernels into the SISR task to enhance frequency, then LSKs are decomposed into 1D kernels to reduce the number of parameters (Fig.~\ref{fig:cnn-p}).

    \subsection{Learnable separable kernel}\label{sec:learnable-seprable-kernel}

    LSKs are essentially rank-one $n \times n$ learnable matrices, which can be decomposed into two multiplicative $n \times 1$ and $ 1 \times n$ matrices. 
    
    As shown in Fig.~\ref{fig:cnn-p}, for square kernels, the $i^{th}$ feature map $f_{m}^{i}$ of the $m^{th}$ layer is generated by activating additive convolved upper layer feature maps $\{f_{m-1}\}$. This process can be expressed as
    \begin{equation}
        \label{eq:f_m}
        f_{m}^{i}=\sigma(\sum_{t=1}^{n_{f_{m-1}}}(f_{m-1}^{t}*(k_{t,i})_{n\times n})),
    \end{equation}
    where $f_{m-1}^{t}$ denotes the $t^{th}$ feature maps of the upper layer, $k_{t,i}$ denotes the square kernels, and $\sigma$ denotes the activation function.
    

    For LSKs, we decompose the LSKs into two layers orthogonal 1D kernels. At this point, an extra layer of feature maps will be generated:
    \begin{equation}
        \label{eq:f_e}
        f_{e}^{j}=\sum_{t=1}^{n_{f_{m-1}}} (f_{m-1}^{t} *\left(k_{t,j}\right)_{n \times 1}),
    \end{equation}
    where $k_{t,j}$ denotes the kernel between two feature maps: $f_{e}^{j}$ and $f_{m-1}^{t}$.
    Note that we get $f_{e}$ without activation, so we can merge the linear operations.
    At present, as shown in right side of the Fig.~\ref{fig:cnn-p}, the $f_{m}^{i}$ can be denoted by
    \begin{equation}
        \label{eq:f_m_s}
        \begin{aligned}
            f_{m}^{i}&=\sigma(\sum_{j=1}^{n_{f_{e}}} f_{e}^{j} *\left(k_{j,i}\right)_{1 \times n})\\
            &=\sigma(\sum_{j=1}^{n_{f_{e}}} (\sum_{t=1}^{n_{f_{m-1}}} f_{m-1}^{t} *\left(k_{t,j}\right)_{n \times 1}) *\left(k_{j,i}\right)_{1 \times n}) \\
            & =\sigma(\sum_{j=1}^{n_{f_{e}}} \sum_{t=1}^{n_{f_{m-1}}} f_{m-1}^{t} *\left(k_{t,j}\right)_{n \times 1} *\left(k_{j,i}\right)_{1 \times n})\\
            & =\sigma(\sum_{t=1}^{n_{f_{m-1}}} (f_{m-1}^{t} *\sum_{j=1}^{n_{f_{e}}}(\left(k_{t,j}\right)_{n \times 1} *\left(k_{j,i}\right)_{1 \times n}))),\\
        \end{aligned}
    \end{equation}
    where $k_{j,i}$ denotes the kernels between $f_{e}^{j}$ and $f_{m}^{i}$.
    The result of multiplying $k_{t,j}$ and $k_{j,i}$ can be represented by the LSK $k_s$.

    Comparing Equation~\eqref{eq:f_m} and Equation~\eqref{eq:f_m_s}, our work is like replacing square kernels with
    the sum of the LSKs:
    \begin{equation}
        \label{eq:change}
        k \Rightarrow \sum k_s.
    \end{equation}
    Base on the characteristics of the LSKs and the specificity of SISR task, the model with LSKs would have inherently capable to increase frequency.
    

    \subsection{LSKs for frequency enhancement}\label{subsec:frequency-enhanced-separable-kernels}
    A rank-1 matrix can enhance the frequency. For example, in the field of image processing, non-learnable separable kernels such as Sobel are often used for edge feature extraction or edge feature enhancement. The blurred LR images miss much high-frequency information after down-sampling from Ground Truth (GT) images, we can achieve better reconstruction results by enhancing the frequency. Based on the characteristics (rank-1) of the LSKs and the specificity of the SISR task, the LSKs are suitable for SISR.

    \subsection{Complexity analysis}\label{subsec:complexity-analysis}

    \subsubsection{Space complexity of the LSKs.}\label{subsubsec:lightweight-separable-kernels}
    The most direct way to compress the model is to reduce the parameters of the kernels. For CNNs, the feature maps of the upper layer are convoluted with kernels, and then the next feature maps are generated by adding the results after activation.

    As shown in left side of the Fig.~\ref{fig:cnn-p}, between two layers, the number of parameters of the normal kernel can be
    denoted as
    \begin{equation}
        \label{eq:e2}
        P_N = n_{f_{m-1}} \times n_{k} \times n_{k} \times n_{f_{m}},
    \end{equation}
    where $n_{f_{m-1}}$ denotes the number of feature maps of the upper layer, $n_{k} \times n_{k}$
    denotes the number of a square kernel and $n_{f_{m}}$ denotes the number of feature maps
    of the next layer.

    As shown in right side of the Fig.~\ref{fig:cnn-p}, we use LSKs instead of normal kernels, the number of parameters
    of which can be denoted as
    \begin{equation}
        \label{eq:e3}
        P_S = n_{f_{m-1}} \times n_{k} \times n_{f_e}+n_{f_e} \times n_{k} \times n_{f_{m}},
    \end{equation}
    where the $n_{f_e}$ represents the number of extra feature maps.
    The former part of Equation~\eqref{eq:e3} represents the number of parameters of 1D kernels between the upper layer and the extra layer,
    and the latter part represents the number of parameters of 1D kernels between extra layer and next layer.

    In general, we can make empirical assumption: $n_{f_{m-1}}=n_{f_e}=n_{f_{m}}$, then we would get:
    \begin{equation}
        \label{eq:eta}
        \begin{aligned}
            \eta &= \frac{P_S}{P_N} = \frac{n_{f_{m-1}} \times n_{k} \times n_{f_e}+n_{f_e} \times n_{k} \times n_{f_{m}}}
            {n_{f_{m-1}} \times n_{k} \times n_{k} \times n_{f_{m}}}\\
            &=\frac{2}{n_{k}}.
        \end{aligned}
    \end{equation}
    In this assumption, we know that $P_S$ is $\eta$ times of ${P_N}$, and the parameters become $\frac{2}{n_{k}}$ times original. Normally, the ${n_{k}} > 3$ and the parameters will be significantly reduced.


    \subsubsection{Time complexity of the LSKs.}\label{subsubsec:efficient-separable-kernels}
    Floating point operations (Flops) are used to measure the complexity of the methods. To better represent the computational complexity, we represent multiplication and addition quantity separately through complex values because multiplication consumes more resources than addition. Therefore, the number of the Flops of once convolution operation ($F_{k\times k}$) can be expressed as
    \begin{equation}
        \label{eq:one}
        F_{k\times k}=n_k\times n_k + (n_k\times n_k-1)i,
    \end{equation}
    where the real part of $F_{k\times k}$ represents the multiplication quantity, and the
    imaginary part represents the addition quantity.
    Then the Flops required of getting the $m$-th layer can be denoted by
    \begin{equation}
        \label{eq:Fm}
        \begin{aligned}
            F_m&=(n_{f_{m-1}}whF_{k\times k} + n_{f_{m-1}}i)\times n_{f_m}\\
            &=Awh n_k^2+A(wh(n_k^2-1)+1)i,
        \end{aligned}
    \end{equation}
    where $w$ and $h$ represent the width and height of the last layer feature map, $A=n_{f_{m-1}} \times n_{f_m}$.

    After we replace the square kernels with LSKs, the needed Flops for the extra layer ($F_e$) can be denoted by
    \begin{equation}
        \label{eq:Fe}
        \begin{aligned}
            F_e&=(n_{f_{m-1}} whF_{k\times 1} + n_{f_{m-1}}i)\times n_{f_e}\\
            &=Bwh n_k+B(wh(n_k-1)+1)i,
        \end{aligned}
    \end{equation}
    where $B=n_{f_{m-1}} \times n_{f_e}$.
    The corresponding needed Flops of the $m$-th layer can be denoted by
    \begin{equation}
        \label{eq:Fsm}
        \begin{aligned}
            F_m'&=(n_{f_{e}}whF_{1\times k} + n_{f_{e}}i)\times n_{f_m}\\
            &=Cwh n_k+C(wh(n_k-1)+1)i,
        \end{aligned}
    \end{equation}
    where $C=n_{f_{e}} \times n_{f_m}$.
    Thus, the all needed Flops ($F_a$) between two layers after decomposition can be denoted by
    \begin{equation}
        \label{eq:Fa}
        \begin{aligned}
            F_a&=F_e+F_m'\\
            &=(B+C)wh n_k + (B+C)(wh(n_k-1)+1)i.
        \end{aligned}
    \end{equation}
    Then, we compare multiplication and addition quantities according to Equation~\ref{eq:Fm} and Equation~\ref{eq:Fa}
    separately.
    Since the $F_m$ and $F_a$ are difficult to compare intuitively, we adopt the empirical assumption:
    $n_{f_{m-1}}=n_{f_{e}}=n_{f_{m}}$, then we get:
    \begin{equation}
        \label{eq:compare}
        \begin{aligned}
            \alpha &= \frac{Real(F_m)}{Real(F_a)} = \frac{Awh n_k^2}{(B+C)wh n_k}\\
            &=\frac{n_k}{2}\\
            \beta  &=\frac{Imag(F_m)}{Imag(F_a)} = \frac{A(wh(n_k^2-1)+1)}{(B+C)(wh(n_k-1)+1)}\\
            &\approx \frac{n_k+1}{2},
        \end{aligned}
    \end{equation}
    where $Real(\cdot)$ and $Imag(\cdot)$ represent operations for extracting real and imaginary parts, respectively,
    and $\alpha$, $\beta$ represent the ratio of multiplication quantity and addition quantity respectively.
    In practice, $n_k \ge3$, so $\alpha\ge1.5$ and $\beta\ge2$.
    Therefore, the quantity of multiplication and addition are both have decreased.


    \section{Experiments}\label{sec:experiments}


    \subsection{Datasets and configuration}\label{subsec:datasets-methods-and-configuration}
    As shown in~\Cref{tab:Config}, some basic configuration details are given. All methods are conducted under the Pytorch framework with an NVIDIA RTX 3080 GPU.
    \begin{table}[htbp]
        \begin{center}
            \caption{Experimental configurations. For ESPCN and S-ESPCN, the learning rate will become $1\times10^{-3}$ when Epochs $=30$, and $1\times10^{-4}$ when Epochs $=80$.}
            \label{tab:Config}
            \scalebox{0.9}{
                \begin{tabular}{|c|c|c|c|c|c|}
                    \hline
                    Training sets & Methods & Epochs & Batch size & Activation   & Init learning rate \\
                    \hline
                    91-images     & SRCNN   & 400    & 16         & ReLU         & $1\times10^{-4}$   \\
                    91-images     & S-SRCNN & 400    & 16         & ReLU         & $1\times10^{-4}$   \\
                    \hline
                    VOC2012       & ESPCN   & 100    & 64         & tanh+sigmoid & $1\times10^{-2}$   \\
                    VOC2012       & S-ESPCN & 100    & 64         & tanh+sigmoid & $1\times10^{-2}$   \\
                    \hline
                    291-images    & VDSR    & 80     & 16         & ReLU         & 0.1                \\
                    291-images    & S-VDSR  & 80     & 16         & ReLU         & 0.1                \\
                    \hline
                \end{tabular}
            }
        \end{center}

    \end{table}
    Next, we give some specific implementation details for different baseline methods. It should be noted that the number of feature maps in the first layer and last layer is 1. According to Equation~\eqref{eq:e3}, replacing the square kernel with LSKs will increase the model parameters when $n_{f_{m-1}}=1$ or $n_{f_m}=1$. So we replace the normal kernel with LSKs in other layers in our experiments

    \textbf{SRCNN with LSKs}. Referring to SRCNN~\cite{SRCNN}, we use 91-images~\cite{91-images} as the training set and Set5~\cite{Set5} as validation set. We use the best performance network SRCNN(9-5-5) as the baseline, which represents that the network has 3 layers and the size of the corresponding kernel are $9\times 9$, $5\times 5$, and $5\times 5$. And the number of corresponding feature maps is $n_1=64$ and $n_2=32$. We replace the $2^{th}$ layer kernels of the SRCNN(9-5-5) with LSKs and then we get S-SRCNN(9-s5-5), where $\text{s}5$ denotes we have replaced the square kernels into LSKs. We decompose the LSKs into two 1D kernels in the training and deployment stages.


    \textbf{ESPCN with LSKs}.
    Following ESPCN~\cite{ESPCN}, we train and validate datasets from VOC2012~\cite{voc2012} for ESPCN. We use the best performance network ESPCN(5-3-3) as the baseline which represents that the network has 3 layers and the size of the corresponding kernel are $5\times 5$, $3\times 3$, and $3\times 3$. The number of corresponding feature maps is $n_1=64$ and $n_2=32$. Similarly, we use S-ESPCN(5-s3-5) to denote the separable version.

    \textbf{VDSR with LSKs}.
    Referring to~\cite{VDSR}, we also use 291-images (91 images from 91-images~\cite{91-images} and 200 images from BSD~\cite{BSD}). Since the ResNet\cite{ResNet} allows deeper networks, we conduct experiments with VDSR(B$^\text{N}$) where N denotes the number of blocks(B). The block is composed of the activation function and kernels with size $3\times 3$. And S-VDSR(B$^\text{N}$) denotes the separable versions.

    The unique difference between the separable version methods and the normal version methods is that the original kernels are replaced by LSKs. And the number of extra layer feature maps of all separable versions above is $n_{f_e}=n_{f_m}$.

    \subsection{Benchmark results}\label{subsec:benchmark-results}
    We use several benchmark architectures on Set5~\cite{Set5}, Set14~\cite{Set14}, BSD100~\cite{BSD}, Urban100~\cite{Urban100}, and M109\cite{M109} to evaluate the capability of the LSKs, and then further discuss LSKs in terms of frequency enhancement. As in the previous works, bicubic is used when conducting experiments and then PSNR and SSIM~\cite{SSIM} are used as the performance metrics.

    In the SISR framework, most of the work is based on three methods including SRCNN, ESPCN, and VDSR. Therefore, we use LSKs to replace the normal kernels in these three basic methods. We compare the SISR methods, including SRCNN, ESPCN, and VDSR, with their separable versions on $\times2$, $\times3$, and $\times4$ upscaling tasks. The performance comparison of different normal SISR methods and separable versions on five benchmark datasets is shown in Table~\ref{tab:results}.
    
    \begin{table}[htbp]
        \begin{center}
            \caption{Performance comparison of normal methods and their separable versions on five benchmarks. PSNR(dB)/SSIM on Y channel are reported on each dataset. The \textbf{bold} data indicate that the separable version is better than the normal version, while the \underline{underline} is the opposite.}
            \label{tab:results}
            \scalebox{0.8}{
                \begin{tabular}{|c|l|c|c|c|c|c|}
                    \hline
                    Scale & Method                 & Set5                                 & Set14                                & BSD100                               & Urban100                             & M109                                 \\
                    \hline
                    \multirow{10}{*}{$\times$2}
                    & SRCNN\cite{SRCNN}      & 37.06/\underline{0.9401}             & 34.50/\underline{0.8772}             & 35.15/\underline{0.8001}              & 31.35/\underline{0.8887}            & 32.46/\underline{0.9145}\\
                    & \textbf{S-SRCNN}          & \textbf{37.38}/0.9379                   & \textbf{35.14}/0.8748                   & \textbf{35.82}/0.7983                   & \textbf{32.23}/0.8803              & \textbf{33.17}/0.9123 \\
                    \cline{2-7}
                    & ESPCN\cite{ESPCN}      & \underline{33.68}/\underline{0.9190} & \underline{29.11}/\underline{0.8610} & \underline{26.08}/\underline{0.7922}  & 26.67/\underline{0.8554}            & 27.47/\underline{0.9031} \\
                    & \textbf{S-ESPCN}          & 33.61/0.9161                         & 29.05/0.8583                         & 26.07/0.7903                         & \textbf{26.69}/0.8543                   & \textbf{27.51}/0.9014                   \\
                    \cline{2-7}
                    & VDSR(B$^1$)\cite{VDSR} & 36.41/0.9561                         & \underline{32.12}/\underline{0.9103} & \underline{31.27}/\underline{0.8897}  & 29.22/0.8920                       & \underline{35.00}/\underline{0.9612}          \\
                    & \textbf{S-VDSR(B$^1$)}    & \textbf{36.41/0.9564}                   & 32.03/0.9100                         & 31.23/0.8896                         & \textbf{29.23/0.8927}               & 34.88/0.9608                   \\
                    \cdashline{2-7}
                    & VDSR(B$^2$)\cite{VDSR} & \underline{36.81}/\underline{0.9584} & \underline{32.41}/\underline{0.9134} & \underline{31.50}/\underline{0.8933}  & \underline{29.77}/0.8992            & \underline{35.75}/\underline{0.9650}          \\
                    & \textbf{S-VDSR(B$^2$)}    & 36.72/0.9579                         & 32.34/0.9128                         & 31.48/0.8924                         & 29.70/\textbf{0.8993}                   & 35.72/0.9647                         \\
                    \cdashline{2-7}
                    & VDSR(B$^3$)\cite{VDSR} & \underline{37.03}/\underline{0.9596} & \underline{32.56}/\underline{0.9151} & 31.58/\underline{0.8952}              & \underline{30.06}/\underline{0.9049}            & \underline{36.09}/\underline{0.9674}          \\
                    & \textbf{S-VDSR(B$^3$)}    & 37.00/0.9594                         & 32.50/0.9149                         & \textbf{31.58}/0.8950                   & 29.91/0.9035                         & 35.88/0.9666                         \\
                    \hline
                    \multirow{10}*{$\times$3}
                    & SRCNN\cite{SRCNN}      & 34.43/\underline{0.8655}             & 33.08/\underline{0.7718}             & 34.43/\underline{0.7662}              & 30.63/\underline{0.7536}            & 30.24/\underline{0.8653}          \\
                    & \textbf{S-SRCNN}          & \textbf{34.64}/0.8616                   & \textbf{33.52}/0.7692                   & \textbf {34.90}/0.7635                  & \textbf{31.35}/0.7478               & \textbf{30.92}/0.8580 \\
                    \cline{2-7}
                    & ESPCN\cite{ESPCN}      & 28.33/\underline{0.8375}             & 25.72/\underline{0.7536}             & \underline{25.84}/\underline{0.7397}  & 22.78/\underline{0.7240}            & 25.80/0.8390                   \\
                    & \textbf{S-ESPCN}          & \textbf{28.38}/0.8372                   & \textbf{25.73}/0.7530                   & 25.83/0.7377                         & \textbf{22.80}/0.7234                   & \textbf{25.81/0.8391} \\
                    \cline{2-7}
                    & VDSR(B$^1$)\cite{VDSR} & 32.12/\underline{0.9056}             & 28.62/0.8225                         & \underline{28.16}/0.7810              & \underline{25.72}/\underline{0.7878}    & 29.23/\underline{0.8885}          \\
                    & \textbf{S-VDSR(B$^1$)}    & \textbf{32.12}/0.9038                   & \textbf{28.62/0.8223}                   & 28.14/\textbf{0.7820}                   & 25.70/0.7853 & \textbf{29.36}/0.8884 \\
                    \cdashline{2-7}
                    & VDSR(B$^2$)\cite{VDSR} & 32.56/0.9117                         & 28.91/\underline{0.8309}             & 28.36/\underline{0.7899}              & 26.06/0.7999                                 & 30.03/0.9027                   \\
                    & \textbf{S-VDSR(B$^2$)}    & \textbf{32.56/0.9119}                   & \textbf{28.91}/0.8301                   & \textbf{28.36}/0.7883                   & \textbf{26.08/0.8002}                            & \textbf{30.06/0.9037} \\
                    \cdashline{2-7}
                    & VDSR(B$^3$)\cite{VDSR} & \underline{32.81}/0.9154             & \underline{29.08}/0.8338             & \underline{28.48}/0.7919              & \underline{26.28}/\underline{0.8077}            & \underline{30.55}/\underline{0.9111}          \\
                    & \textbf{S-VDSR(B$^3$)}    & 32.78/\textbf{0.9155}                   & 29.05/\textbf{0.8340}                   & 28.45/\textbf{0.7930}                   & 26.23/0.8062                                        & 30.42/0.9088                   \\
                    \hline
                    \multirow{10}*{$\times$4}
                    & SRCNN\cite{SRCNN}      & 32.78/\underline{0.8353}             & 32.36/\underline{0.7172}             & 34.18/\underline{0.6824}              & 30.29/\underline{0.7068}            & 29.47/\underline{0.7720}          \\
                    & \textbf{S-SRCNN}          & \textbf{33.26}/0.8297                   & \textbf{32.97}/0.7126                   & \textbf{34.82}/0.6792                   & \textbf{31.13}/0.6977                 & \textbf{30.38}/0.7646 \\
                    \cline{2-7}
                    & ESPCN\cite{ESPCN}      & 27.96/\underline{0.7989}             & 24.92/0.6930                         & 24.52/\underline{0.6590}              & 22.36/0.6688                     & 22.84/0.7513                   \\
                    & \textbf{S-ESPCN}          & \textbf{28.06}/0.7985                   & \textbf{25.01/0.6939}                   & \textbf{24.55}/0.6582                   & \textbf{22.44/0.6719}                   & \textbf{22.84/0.7544} \\
                    \cline{2-7}
                    & VDSR(B$^1$)\cite{VDSR} & 29.67/0.8461                         & 26.73/0.7483                         & 26.69/0.6981                         & 23.91/0.6991                         & 26.31/0.8131                         \\
                    & \textbf{S-VDSR(B$^1$)}    & \textbf{29.83/0.8499}                   & \textbf{26.80/0.7498}                   & \textbf{26.70/0.6986}                   & \textbf{24.05/0.7014}                      & \textbf{26.44/0.8170} \\
                    \cdashline{2-7}
                    & VDSR(B$^2$)\cite{VDSR} & 30.09/0.8564                         & 27.03/\underline{0.7577}             & 26.86/\underline{0.7068}           & 24.27/0.7141                     & 27.06/0.8313                   \\
                    & \textbf{S-VDSR(B$^2$)}    & \textbf{30.12/0.8578}                   & \textbf{27.08}/0.7562                   & \textbf{26.86}/0.7039                   & \textbf{24.27/0.7144}                      & \textbf{27.12/0.8334} \\
                    \cdashline{2-7}
                    & VDSR(B$^3$)\cite{VDSR} & \underline{30.34}/\underline{0.8631} & \underline{27.12}/\underline{0.7620} & \underline{26.92}/\underline{0.7099}      & \underline{24.42}/\underline{0.7222}            & \underline{27.30}/\underline{0.8406}          \\
                    & \textbf{S-VDSR(B$^3$)}    & 30.28/0.8603                         & 27.09/0.7589                         & 26.91/0.7078                         & 24.36/0.7171                         & 27.28/0.8378                         \\
                    \hline
                \end{tabular}
            }
        \end{center}
        
    \end{table}

    It can be seen that S-SRCNN performs better than SRCNN on most scales in PSNR. Although the normal version models perform better on some datasets in PSNR, this superiority hardly exceeds \textbf{0.1dB}. This indicates that the separable version model is comparable to the normal model.

    \subsection{Discuss}\label{subsec:discuss}
    In this section, we would make a qualitative and quantitative discussion about the LSKs through benchmark results (~\Cref{tab:results}).

    \subsubsection{Parameters and Flops.}
    The parameters reflect the size of the model, and Flops can be used to measure the complexity of the model. As shown in Table~\ref{tab:p_f}, the parameters and Flops have been greatly reduced with LSKs. For example, the number of parameters and Flops of the S-SRCNN has decreased by more than 60\% compared with the SRCNN. Consistent with the analysis in section~\ref{sec:methodology}, the larger the size of the kernel, the more the amount of parameters (~\Cref{subsubsec:lightweight-separable-kernels}) and Flops (~\Cref{subsubsec:efficient-separable-kernels}) would be reduced.
    \begin{table}[htbp]
        \begin{center}
            \caption{The comparison of parameters and Flops between separable version models and their normal version.
            We also list the decline in parameter and Flops of the separable version models compared to the normal version.
            The number of Flops is calculated under the setting of upscaling image to $512 \times 512$ resolution
            on $\times2$, $\times3$ and $\times4$ tasks.}
            \label{tab:p_f}
            \scalebox{1}{
                \begin{tabular}{|l|c|c|c|c|}
                    \hline
                    Metrics                & Parameters(K)$\downarrow$ & Decline(\%)     & Flops(G)$\downarrow$ & Decline(\%)        \\
                    \hline
                    SRCNN\cite{SRCNN}      & 57.23                     & \multirow{2}*{\textbf{62.48}} & 15.02                & \multirow{2}*{\textbf{62.45}} \\
                    \textbf{S-SRCNN}          & \textbf{21.47}               &                            & \textbf{5.64}           &                            \\
                    \hline
                    ESPCN\cite{ESPCN}      & 21.28                     & \multirow{2}*{\textbf{43.14}} & 5.58                 & \multirow{2}*{\textbf{43.19}} \\
                    \textbf{S-ESPCN}          & \textbf{12.10}               &                            & \textbf{3.17}           &                            \\
                    \hline
                    VDSR(B$^1$)\cite{VDSR} & 38.08                     & \multirow{2}*{\textbf{32.09}} & 10.02                & \multirow{2}*{\textbf{32.04}} \\
                    \textbf{S-VDSR(B$^1$)}    & \textbf{25.86}               &                            & \textbf{6.81}           &                            \\
                    \cdashline{1-5}
                    VDSR(B$^2$)\cite{VDSR} & 75.01                     & \multirow{2}*{\textbf{32.60}} & 19.71                & \multirow{2}*{\textbf{34.04}} \\
                    \textbf{S-VDSR(B$^2$)}    & \textbf{50.56}               &                            & \textbf{13.30}          &                            \\
                    \cdashline{1-5}
                    VDSR(B$^3$)\cite{VDSR} & 111.9                     & \multirow{2}*{\textbf{32.74}} & 29.41                & \multirow{2}*{\textbf{35.12}} \\
                    \textbf{S-VDSR(B$^3$)}    & \textbf{75.26}               &                            & \textbf{19.80}          &                            \\

                    \hline
                \end{tabular}
            }
        \end{center}
        
    \end{table}
    

    \subsubsection{PSNR and SSIM.}
\begin{figure}[htbp]
        \begin{center}
            \includegraphics[width=0.5\linewidth]{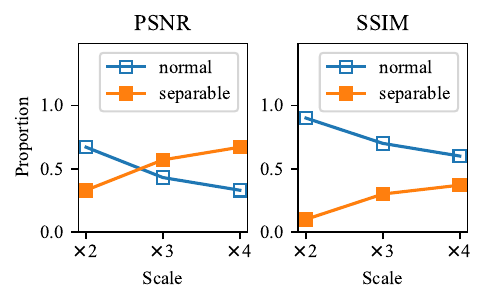}
        \end{center}
        \caption{The line graph of the proportion of better versions under different scale factors in PSNR and SSIM.
        The trend of the separable version methods is gradually improving with the increase of scale factor. It is worth noting that although the normal version sometimes outperforms the separable version in terms of PSNR, this advantage does not exceed 0.1 dB.}
        \label{fig:psnr_ssim}
    \end{figure}

    To analyze the performance trend of separable version methods with the increase of the scale. We count the proportion (Table~\ref{tab:PSNR and SSIM}) of the methods that perform better than their competitors on PSNR and SSIM respectively. Then, we use a line graph (Fig.~\ref{fig:psnr_ssim}) to show the proportion of different scale factors through Table~\ref{tab:PSNR and SSIM}. As shown in Fig.~\ref{fig:psnr_ssim}, the proportion of separable version methods gradually increased with the increase of scale factor, whether in PNSR or SSIM.

    \begin{table}[htbp]
        \begin{center}
            \caption{The proportion of the methods that perform better than their competitors on PSNR and SSIM respectively.}
            \label{tab:PSNR and SSIM}
            \scalebox{1}{
                \begin{tabular}{|c|c|c|c|c|}
                    \hline
                    Metrics & Method    & $\times2$ & $\times3$ & $\times4$ \\
                    \hline
                    \multirow{2}*{PSNR}
                    & normal    & 0.67      & 0.43      & 0.33      \\
                    & separable & 0.33      & 0.57      & 0.67      \\
                    \hline
                    \multirow{2}*{SSIM}
                    & normal    & 0.90      & 0.70      & 0.63      \\
                    & separable & 0.10      & 0.30      & 0.37      \\
                    \hline
                \end{tabular}
            }
        \end{center}
    
    \end{table}

    The LSKs are more suitable for larger scale factors scenarios, where vast high-frequency information is lost. With the increase of scale factor, more details or frequency information is lost in the LR image. As we analyze in subsection~\ref{subsec:frequency-enhanced-separable-kernels}, the LSKs can enhance the information of frequency.

    \subsection{Interpretability}\label{subsec:visualization-for-learned-kernels}
\begin{figure}[htbp]
        \begin{center}
            \includegraphics[width=1\linewidth]{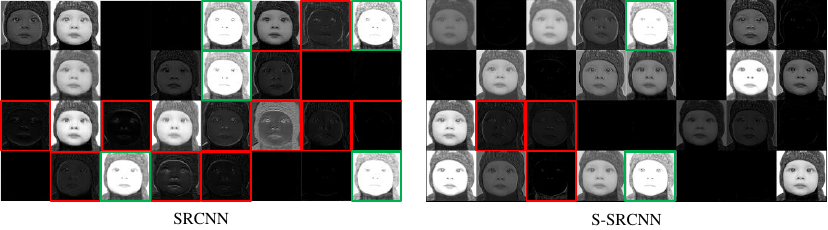}
        \end{center}
        \caption{The comparison of the last layer feature maps between the SRCNN and S-SRCNN with scale factor $\times 4$.
        Feature maps related to frequency and lacking details are marked with red boxes and green boxes, separately.}
        \label{fig:v_feature}
    \end{figure}

    One of the important works of SISR is to reconstruct the details of LR images. The most common way to recover detail is to enhance frequency. The HR image is generated by the last feature maps, so we compare the last layer feature maps between the SRCNN and S-SRCNN. Among them, the feature maps of the SRCNN are generated by square kernels, and the feature maps of the S-SRCNN are generated by LSKs. As shown in Fig.~\ref{fig:v_feature}, we visualize the feature maps and mark the frequency-related with red boxes and detail-lacking with green boxes.
    
    In the last layer of feature maps of the SRCNN, we can see that the boundary between frequency-related feature maps and detail-lacking feature maps is very clear. Each type of feature map has its contribution. Compared to SRCNN, there are fewer frequency-related and detail-lacking feature maps in S-SRCNN. This phenomenon indicates that most of these factors have been integrated into each feature map. In addition, the S-SRCNN performs better than SRCNN in experiments, therefore, we think that LSKs can directly learn frequency features.

    \section{Conclusion}\label{sec:conclusion}
    In this paper, we propose the LSKs, which can be decomposed into merge-able 1D orthogonal kernels, to substitute the usual square kernels. The LSKs with the ability to enhance frequency are suitable for SISR tasks, especially aiming at large upsampling scales. The comprehensive experiments demonstrate the effectiveness of LSKs, and the parameters and computations of the model are also significantly reduced. Moreover, the visualization results demonstrate the LSKs' ability to enhance image frequency from the interpretable perspective. Future work can apply the plug-and-play LSKs to large models in combination with other auxiliary structures.

\bibliographystyle{splncs04}
\bibliography{mybibfile}

\begin{thebibliography}{10}
\providecommand{\url}[1]{\texttt{#1}}
\providecommand{\urlprefix}{URL }
\providecommand{\doi}[1]{https://doi.org/#1}

\bibitem{gvloss}
Abrahamyan, L., Truong, A.M., Philips, W., Deligiannis, N.: Gradient variance
  loss for structure-enhanced image super-resolution. In: ICASSP 2022-2022 IEEE
  International Conference on Acoustics, Speech and Signal Processing (ICASSP).
  pp. 3219--3223. IEEE (2022)

\bibitem{kernelPruning}
Anwar, S., Hwang, K., Sung, W.: Structured pruning of deep convolutional neural
  networks. ACM Journal on Emerging Technologies in Computing Systems (JETC)
  \textbf{13}(3),  1--18 (2017)

\bibitem{Set5}
Bevilacqua, M., Roumy, A., Guillemot, C., line Alberi~Morel, M.: Low-complexity
  single-image super-resolution based on nonnegative neighbor embedding. In:
  Proceedings of the British Machine Vision Conference. pp. 135.1--135.10. BMVA
  Press (2012)

\bibitem{ACB}
Ding, X., Guo, Y., Ding, G., Han, J.: Acnet: Strengthening the kernel skeletons
  for powerful cnn via asymmetric convolution blocks. In: 2019 IEEE/CVF
  International Conference on Computer Vision (ICCV). pp. 1911--1920 (2019)

\bibitem{DBB}
Ding, X., Zhang, X., Han, J., Ding, G.: Diverse branch block: Building a
  convolution as an inception-like unit. In: 2021 IEEE/CVF Conference on
  Computer Vision and Pattern Recognition (CVPR). pp. 10881--10890 (2021)

\bibitem{RepVGG}
Ding, X., Zhang, X., Ma, N., Han, J., Ding, G., Sun, J.: Repvgg: Making
  vgg-style convnets great again. In: 2021 IEEE/CVF Conference on Computer
  Vision and Pattern Recognition (CVPR). pp. 13728--13737 (2021)

\bibitem{SRCNN}
Dong, C., Loy, C.C., He, K., Tang, X.: Image super-resolution using deep
  convolutional networks. IEEE Transactions on Pattern Analysis and Machine
  Intelligence  \textbf{38}(2),  295--307 (2016)

\bibitem{voc2012}
Everingham, M., Van~Gool, L., Williams, C.K.I., Winn, J., Zisserman, A.: The
  {PASCAL} {V}isual {O}bject {C}lasses {C}hallenge 2012 {(VOC2012)} {R}esults.
  http://www.pascal-network.org/challenges/VOC/voc2012/workshop/index.html

\bibitem{ResNet}
He, K., Zhang, X., Ren, S., Sun, J.: Deep residual learning for image
  recognition. In: Proceedings of the IEEE conference on computer vision and
  pattern recognition. pp. 770--778 (2016)

\bibitem{Urban100}
Huang, J.B., Singh, A., Ahuja, N.: Single image super-resolution from
  transformed self-exemplars. In: 2015 IEEE Conference on Computer Vision and
  Pattern Recognition (CVPR). pp. 5197--5206 (2015)

\bibitem{low_rank1}
Jaderberg, M., Vedaldi, A., Zisserman, A.: Speeding up convolutional neural
  networks with low rank expansions. In: Proceedings of the British Machine
  Vision Conference (2014)

\bibitem{VDSR}
Kim, J., Lee, J.K., Lee, K.M.: Accurate image super-resolution using very deep
  convolutional networks. In: 2016 IEEE Conference on Computer Vision and
  Pattern Recognition (CVPR). pp. 1646--1654 (2016)

\bibitem{medical_sr}
Li, Y., Sixou, B., Peyrin, F.: A review of the deep learning methods for
  medical images super resolution problems. Irbm  \textbf{42}(2),  120--133
  (2021)

\bibitem{BSD}
Martin, D., Fowlkes, C., Tal, D., Malik, J.: A database of human segmented
  natural images and its application to evaluating segmentation algorithms and
  measuring ecological statistics. In: Proceedings Eighth IEEE International
  Conference on Computer Vision. ICCV 2001. vol.~2, pp. 416--423. IEEE (2001)

\bibitem{M109}
Matsui, Y., Ito, K., Aramaki, Y., Fujimoto, A., Ogawa, T., Yamasaki, T.,
  Aizawa, K.: Sketch-based manga retrieval using manga109 dataset. Multimedia
  Tools and Applications  \textbf{76},  21811--21838 (2017)

\bibitem{ESPCN}
Shi, W., Caballero, J., Huszár, F., Totz, J., Aitken, A.P., Bishop, R.,
  Rueckert, D., Wang, Z.: Real-time single image and video super-resolution
  using an efficient sub-pixel convolutional neural network. In: 2016 IEEE
  Conference on Computer Vision and Pattern Recognition (CVPR). pp. 1874--1883
  (2016)

\bibitem{img_restore}
Su, J., Xu, B., Yin, H.: A survey of deep learning approaches to image
  restoration. Neurocomputing  \textbf{487},  46--65 (2022)

\bibitem{google-v3}
Szegedy, C., Vanhoucke, V., Ioffe, S., Shlens, J., Wojna, Z.: Rethinking the
  inception architecture for computer vision. In: Proceedings of the IEEE
  conference on computer vision and pattern recognition. pp. 2818--2826 (2016)

\bibitem{cm1}
Timofte, R., De, V., Gool, L.V.: Anchored neighborhood regression for fast
  example-based super-resolution. In: 2013 IEEE International Conference on
  Computer Vision. pp. 1920--1927 (2013)

\bibitem{cm2}
Timofte, R., De~Smet, V., Van~Gool, L.: A+: Adjusted anchored neighborhood
  regression for fast super-resolution. In: Computer Vision--ACCV 2014: 12th
  Asian Conference on Computer Vision, Singapore, Singapore, November 1-5,
  2014, Revised Selected Papers, Part IV 12. pp. 111--126. Springer (2015)

\bibitem{EFDN}
Wang, Y.: Edge-enhanced feature distillation network for efficient
  super-resolution. In: 2022 IEEE/CVF Conference on Computer Vision and Pattern
  Recognition Workshops (CVPRW). pp. 776--784 (2022)

\bibitem{sr_survey}
Wang, Z., Chen, J., Hoi, S.C.H.: Deep learning for image super-resolution: A
  survey. IEEE Transactions on Pattern Analysis and Machine Intelligence
  \textbf{43}(10),  3365--3387 (2021)

\bibitem{SSIM}
Wang, Z., Bovik, A., Sheikh, H., Simoncelli, E.: Image quality assessment: from
  error visibility to structural similarity. IEEE Transactions on Image
  Processing  \textbf{13}(4),  600--612 (2004)

\bibitem{FAD}
Xie, W., Song, D., Xu, C., Xu, C., Zhang, H., Wang, Y.: Learning
  frequency-aware dynamic network for efficient super-resolution. In: 2021
  IEEE/CVF International Conference on Computer Vision (ICCV). pp. 4288--4297
  (2021)

\bibitem{91-images}
Yang, J., Wright, J., Huang, T.S., Ma, Y.: Image super-resolution via sparse
  representation. IEEE Transactions on Image Processing  \textbf{19}(11),
  2861--2873 (2010)

\bibitem{Set14}
Zeyde, R., Elad, M., Protter, M.: On single image scale-up using
  sparse-representations. In: Curves and Surfaces: 7th International
  Conference, Avignon, France, June 24-30, 2010, Revised Selected Papers 7. pp.
  711--730. Springer (2012)

\bibitem{pruning_layer}
Zhan, Z., Gong, Y., Zhao, P., Yuan, G., Niu, W., Wu, Y., Zhang, T., Jayaweera,
  M., Kaeli, D., Ren, B., Lin, X., Wang, Y.: Achieving on-mobile real-time
  super-resolution with neural architecture and pruning search. In: 2021
  IEEE/CVF International Conference on Computer Vision (ICCV). pp. 4801--4811
  (2021)

\bibitem{low_rank2}
Zhang, X., Zou, J., He, K., Sun, J.: Accelerating very deep convolutional
  networks for classification and detection. IEEE Transactions on Pattern
  Analysis and Machine Intelligence  \textbf{38}(10),  1943--1955 (2016)

\bibitem{ECB}
Zhang, X., Zeng, H., Zhang, L.: Edge-oriented convolution block for real-time
  super resolution on mobile devices. In: Proceedings of the 29th ACM
  International Conference on Multimedia. pp. 4034--4043 (2021)

\end{thebibliography}

\end{document}